\documentclass[10pt,twocolumn,letterpaper]{article}

\usepackage{cvm}
\usepackage{times}
\usepackage{epsfig}
\usepackage{graphicx}
\usepackage{amsmath}
\usepackage{amssymb}

\usepackage[pagebackref=true,breaklinks=true,letterpaper=true,colorlinks,bookmarks=false]{hyperref}

\newcommand{\keywords}[1]{{\bf \emph{Keywords: #1}}}

\cvmfinalcopy

\def\cvmPaperID{****} 

\ifcvmfinal\fi
\begin{document}

\title{ AEANet: Affinity-Enhanced Attentional Networks for Arbitrary Style Transfer }

\author{ Gen Li  \\
Guilin University Of Electronic Technology\\
Guilin, China\\
{\tt\small alroy@mails.guet.edu.cn}
\and
Xianqiu Zheng\\
Guilin University Of Electronic Technology\\
Guilin, China\\
{\tt\small 22031102009@mails.guet.edu.cn}
\and
Yujian Li\\
Guilin University Of Electronic Technology\\
Guilin, China\\
{\tt\small liyujian@guet.edu.cn}
}

\maketitle

\begin{abstract}
Arbitrary artistic style transfer is a research area that combines rational academic study with emotive artistic creation. It aims to create a new image from a content image according to a target artistic style, maintaining the content's textural structural information while incorporating the artistic characteristics of the style image. However, existing style transfer methods often significantly damage the texture lines of the content image during the style transformation. To address these issues, we propose affinity-enhanced attentional network, which include the content affinity-enhanced attention (CAEA) module, the style affinity-enhanced attention (SAEA) module, and the hybrid attention (HA) module. The CAEA and SAEA modules first use attention to enhance content and style representations, followed by a detail enhanced (DE) module to reinforce detail features. The hybrid attention module adjusts the style feature distribution based on the content feature distribution. We also introduce the local dissimilarity loss based on affinity attention, which better preserves the affinity with content and style images. Experiments demonstrate that our work achieves better results in arbitrary style transfer than other state-of-the-art methods.
\end{abstract}


\section{Introduction}

Scientific research is rational and rigorous, artistic creation is emotional and dynamic, and artistic style migration is a study that can perfectly balance scientific research and artistic creation. This study aims to disassemble the color and texture 
\begin{figure}[th]
\begin{center}
    \centering
    \includegraphics[width=0.5\textwidth]{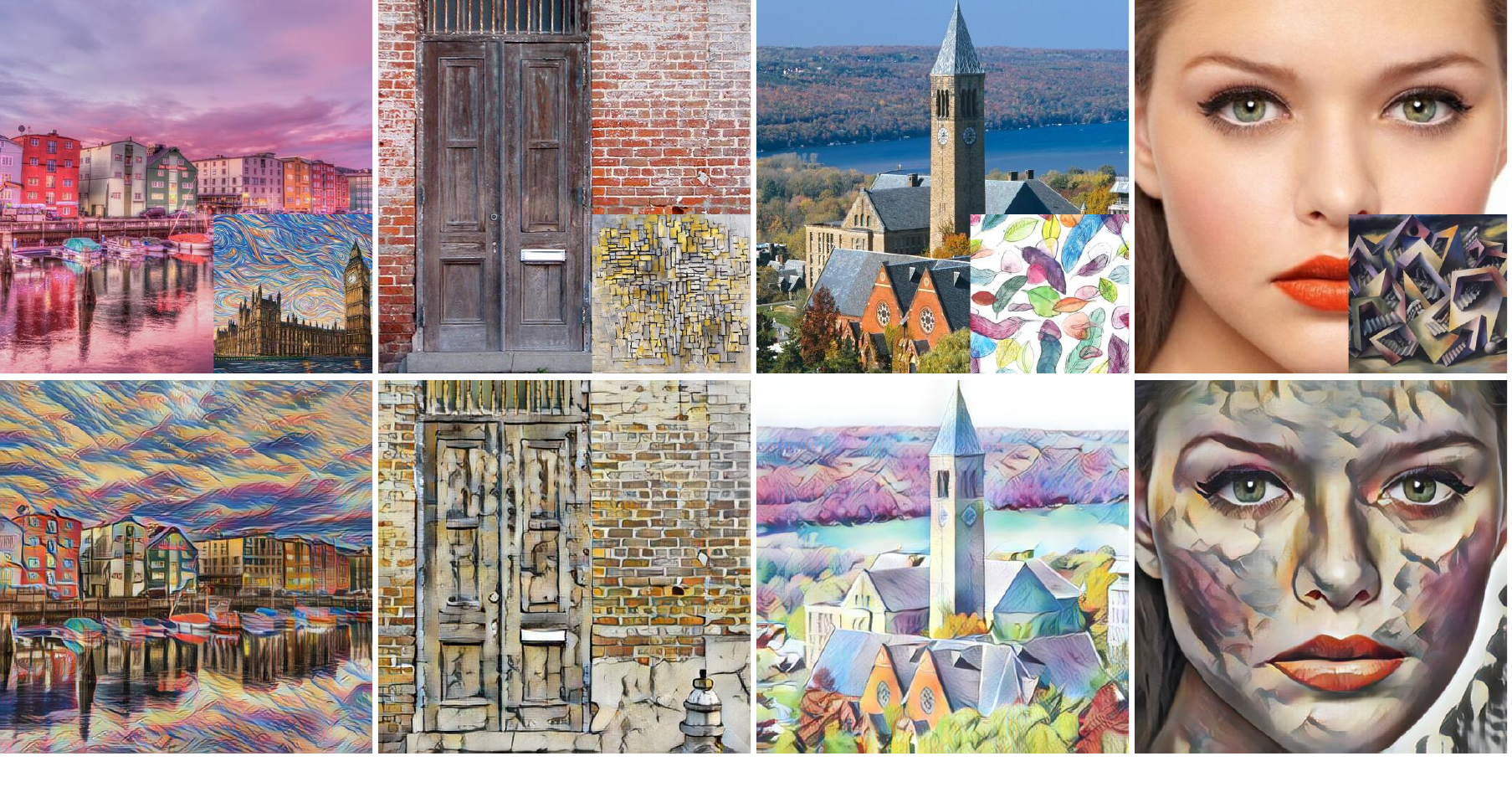}
\end{center}
   \caption{Utilizing our arbitrary style transfer model, the generated images exhibit substantial enhancements in visual quality and a marked improvement in content affinity. }
\label{fig:long}
\end{figure}
features in a stylized image and migrate the features to a given content image to achieve image style conversion while preserving the features of the content image as much as possible under the constraints of style  migration. In recent years, Gatys \etal~\cite{gatys2016image} first used pre-trained deep convolutional neural networks~\cite{2014Very} to separate and recombine images' content and style features. They innovatively used the Gram matrix to quantify the correlations of style features, capturing and transferring the color and texture layout of the target style image. However, the method proposed by Gatys \etal~\cite{gatys2016image} relies on an iterative gradient descent process to optimize the image, resulting in prolonged image generation speeds. 

To expand the application scenarios of style transfer, some researchers have used feedforward network-based methods \cite{2016Perceptual,2019Style,2017StyleBank,li2017diversified,ulyanov2016texture,zhang2018multi,qi2023icdaelst} to accomplish this task efficiently. However, these methods can only achieve the transfer of a specific style. Another group of researchers has been striving to extend the applicability of this research to arbitrary style transfer. \cite{kalischek2021light,li2017universal, huang2017arbitrary,an2021artflow,ho2019flow++,deng2021arbitrary,jing2020dynamic} They focus on aligning global statistics by adjusting the overall distribution of the content image to match the second-order statistics of the style image. 
Huang \etal~\cite{huang2017arbitrary} adjusted the feature statistics (mean and variance) of the content image to match the statistical characteristics of the target style image. Li \etal~\cite{li2017universal}  used whitening and coloring transforms to ensure that the feature covariance of the content image directly matches the statistical characteristics of the target style image. Specifically, the WCT~\cite{li2017universal} method removes the variance influence of content features through a whitening step and then transforms these features into a space that matches the covariance of the style image through a coloring process. This process goes beyond simple variance matching, achieving deeper style embedding. However, the high computational cost of WCT~\cite{li2017universal} limits its efficiency and real-time application.
Correspondingly, local patch-based methods \cite{chen2016fast,gu2018arbitrary,park2019arbitrary,sheng2018avatar,chen2021artistic,deng2020arbitrary,yao2019attention,zhang2019multimodal,huo2021manifold,deng2022stytr2} identify and extract corresponding feature regions in content and style images through local feature matching, employing refined fusion strategies such as local linear transformation or feature map-based region matching. Park \etal~\cite{park2019arbitrary} introduced style-attentional network (SANet) , which use self-attention mechanisms to capture the correlation between style and content images. They effectively and flexibly decorate local style patterns based on the semantic spatial distribution of the content image. Although SANet achieves a good balance between style and content to some extent through its identity loss function and feature embedding, it is challenging to find the optimal style transfer effect or preserve content details during the style transfer process for images with rich content details. 

To address the issue of content image information confusion during style transfer, we propose a new arbitrary style transfer model that achieves high-quality stylized transfer inference while maximizing the retention of content information. This model mimics the human painting process, starting with local enhancements to fine-tune detailed and delicate elements and then moving to the overall depiction of fundamental structures and textures. In our study, we introduce an affinity-enhanced attentional network (AEANet) , which includes content affinity-enhanced attention (CAEA) module, style affinity-enhanced attention (SAEA) module, and hybrid attention (HA) module. The CAEA and SAEA modules extract texture, color, and structural information from the content and style images. The content and style affinity attentions pass deep and shallow features into the agent attention~\cite{han2023agent} and content detail enhanced (DE) modules to enhance content representation. Then, the content and style features undergo adaptive normalization to ensure that their local feature statistics match the statistics of the detail-enhanced features. The attention module synthesizes shallow and deep convolutional neural network features from style and content images, achieving point-by-point feature statistic matching between content and style features. The HA module adjusts style features based on the distribution of content features, better capturing the mapping relationship between content and style features. Based on the AEANet module, we propose a local dissimilarity loss based on affinity-enhanced attention. Specifically, this loss is computed as the difference between the features processed by our content and style affinity attention modules and the features after random permutation. Our main contributions are as follows:
\begin{itemize}
    \item We propose a new AEANet that flexibly matches stylistic semantic information on content features.
    \item We design the CAEA module, SAEA module and HA module to capture local feature information, and to  match global statistical information, and to enhance the detail representation in both content and style images, improving the effect of image stylization significantly.
    \item We present affinity attention based on local dissimilarity loss to achieve better content affinity by preserving the information of content images during style migration.
    \item We conduct many experiments to demonstrate that our method preserves rational details within emotional styles, and our AEANet outperforms current state-of-the-art (SOTA) models both qualitatively and quantitatively.

\end{itemize}

\section{ Related work }

\subsection{Global statistics-based arbitrary style transfer} Global statistics-based methods typically achieve style transfer by adjusting the statistical properties of content features to align with the global statistical characteristics of style features. Li \etal~\cite{li2017universal} employed Whitening and Coloring Transform (WCT) to map content image features into a space where the covariance matches the style image's. Huang \etal~\cite{huang2017arbitrary} adapted the mean and variance of content features using an adaptive instance normalization(AdaIN) layer to match the mean and variance of style features globally. Jing  \etal~\cite{jing2020dynamic} extended Huang's work by introducing a Dynamic Instance Normalization (DIN) module, which uses learnable convolutions to encode the style image and stylize the content image accordingly. An \etal~\cite{an2021artflow} proposed an unbiased style transfer framework that addressed the content leakage issue in existing general style transfer methods (e.g., WCT~\cite{li2017universal} or AdaIN~\cite{huang2017arbitrary}) by achieving unbiased image style transfer through reversible neural flows~\cite{ho2019flow++}. Chen \etal~\cite{chen2023tssat} first establish a global style by aligning the global statistics of content features and style features (mean and variance) and then enrich local style
\begin{figure*}[th]
\begin{center}
    \centering
    \includegraphics[width=1\textwidth]{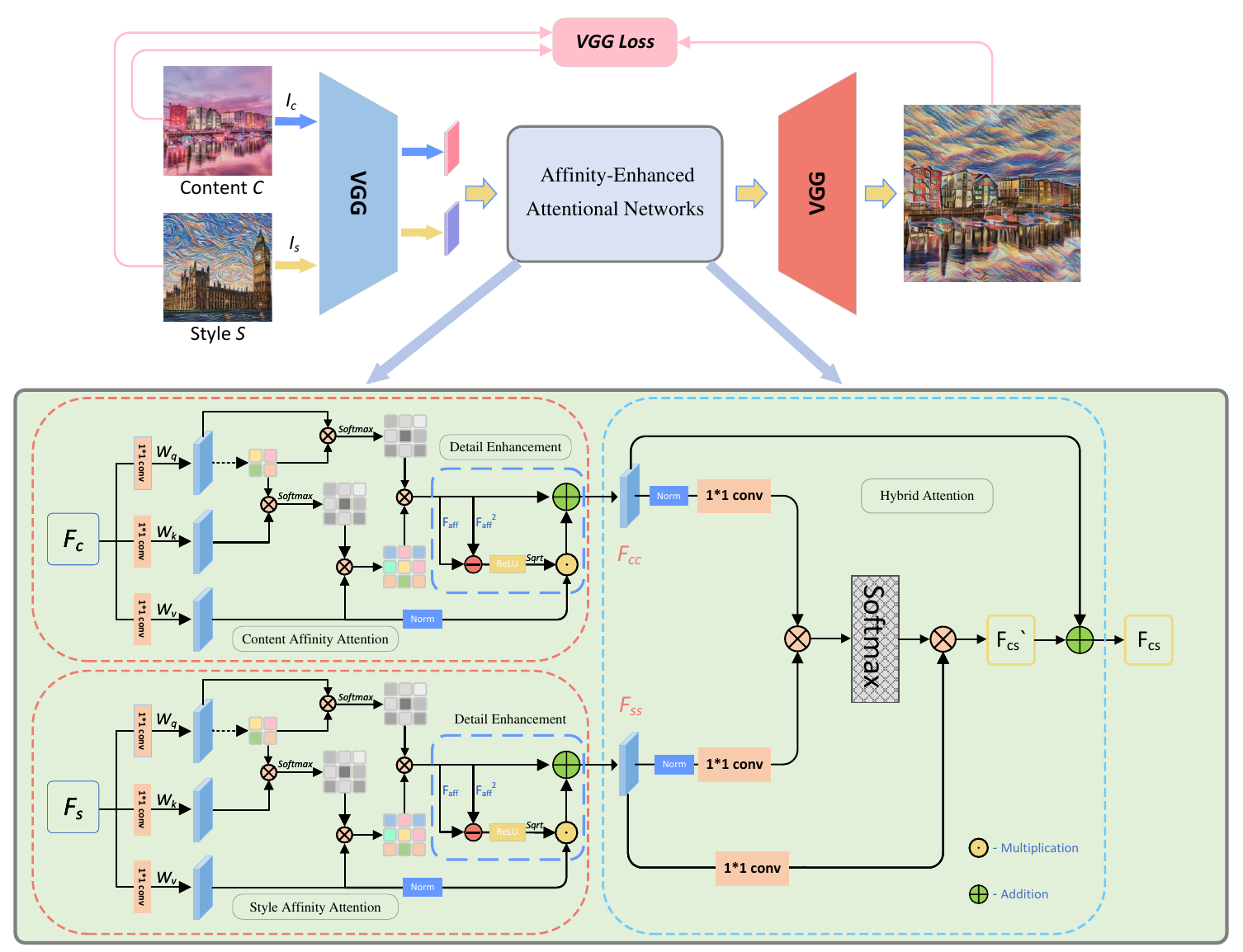}
\end{center}
   \caption{ Overview of our framework. We input the content image \( I_c \) and the style image \( I_s \) into a pre-trained VGG encoder, generating the corresponding features \( F_c \) and \( F_s \). These features \( F_c \) and \( F_s \) are fed into the CAEA and SAEA modules, respectively. The HA module adjusts the style features \( F_{ss} \) based on the distribution of the content features \( F_{cc} \) to obtain the stylized features \( F_{cs} \), which are decoded by a decoder symmetrically designed to the encoder to generate the resulting image \( I_{cs} \).  }
\label{fig:model}
\end{figure*}
details by exchanging local statistical data.
These methods have successfully achieved global distribution alignment between stylized images and the overall style of their corresponding style images. However, they overlook a critical issue: a style image generally encompasses multiple 
stylistic regions, while a content image often contains several distinct semantic areas. Global style transfer needs to fully account for the diversity of local styles and the complexity of multiple semantic features in the content image.

\subsection{Local patch-based arbitrary style transfer}  Local patch-based methods focus on achieving more refined style transfer by paying attention to the color, texture, and regional differences in the local details of images. Chen \etal~\cite{chen2016fast} combined content structures and style textures from a single layer in a pre-trained network, swapping each content patch with the most matching style patch. Park \etal~\cite{park2019arbitrary} introduced the style-attentional network (SANet), which effectively and flexibly decorates local style patterns according to the semantic spatial distribution of the content image while maintaining consistency between global and local style patterns. Building on Park's work, Chen \etal~\cite{chen2021artistic} proposed an artistic style transfer method incorporating internal-external and contrastive learning. By optimizing color and texture patterns with two contrastive loss, they enhanced the harmony and stability of the transformed images. Deng \etal~\cite{deng2020arbitrary} innovatively separated content and style representations through adaptive modules, rearranging the distribution of style features based on content features. Zhang \etal~\cite{zhang2019multimodal} achieved flexible matching of style and content by clustering style image features into sub-style components and matching them with local content features under a graph-cut formulation. Huo \etal~\cite{huo2021manifold} achieved mutual mapping of content and style features by learning a common subspace, then performed style transfer through local semantic alignment. Deng \etal~\cite{deng2022stytr2} proposed a transformer-based style transfer method that uses content-aware positional encoding and a multi-layer transformer decoder to stylize the content sequence based on the style sequence. While these methods excel in capturing more local style features, the generated stylized images often contain unwanted semantic information from the style image and may sometimes deviate from the overall style distribution.

\subsection{Attention mechanism}  Agent attention~\cite{han2023agent} integrates global attention with linear attention by introducing additional agent tokens, thereby combining the expressive power of Softmax attention with the efficiency of linear attention. Our AEANet is related to attention mechanisms utilized in image generation\cite{vaswani2017attention,han2023agent}. The proposed AEANet is designed to learn the features of content images and style images separately.


\section{Method}
\label{sec:3}
Firstly, we briefly introduce the workflow of our model in Section \ref{sec:31}. Secondly, we explain our proposed affinity-enhanced attentional network (AEANet) module in Section \ref{sec:32}. Finally, we describe the loss functions used in AEANet in Section \ref{sec:33}, including our proposed local dissimilarity loss based on affinity-enhanced attention.
\subsection{Overview of AEANet}
\label{sec:31}
Generally, our work involves taking any given style image $ I_s $  and content image $I_c$ and aiming to develop a generative model capable of synthesizing a stylized image $F_{cs}$. This model preserves the texture structure of the target image $I_c$ and learns local and global style features from the style image $I_s$.
To achieve arbitrary artistic style transfer, we propose the affinity-enhanced attentional network (AEANet). The AEANet is composed of a content affinity-enhanced attentional (CAEA) module, style affinity-enhanced attentional (SAEA) module, and  hybrid attention (HA) module. As shown in Figure~\ref{fig:model}, the encoder $E$ and decoder $D$ are symmetrically designed based on the VGG-19 network~\cite{2014Very}, the network inputs are the content image $I_c$ and the style image $I_s$. 

Specifically, the encoder $E$ is a pre-trained VGG-19 network~\cite{2014Very}, since this network was trained on the ImageNet dataset for classification tasks, it is only partially suited for style transfer tasks oriented toward artistic creation. Therefore, its parameters are fixed during subsequent training. The content image $I_c$ and the style image $I_s$ are loaded into $\phi$, which is pre-trained VGG-19 network, to obtain their respective feature maps.

\begin{equation}
F_c = \phi(I_c) , \quad F_s = \phi(I_s) \tag{1},
\end{equation}
after extracting the features $F_c$ and $F_s$, we input both features into the AEANet.

\begin{equation}
F_{cs} = AEANet(F_c, F_s) \quad  \tag{2}
\end{equation}
Specifically, 

\begin{equation*}
F_{cc} = CAEA(F_c)
\end{equation*}
\begin{equation*}
F_{ss} = SAEA(F_s) \tag{3}
\end{equation*}
\begin{equation*}
F_{cs} = HA(F_{cc}, F_{ss})
\end{equation*}
Finally, The feature \( F_{cs} \) is then passed into the decoder \( D \), generating the stylized image \( I_{cs} \).

\begin{align}
I_{cs} = Decoder(F_{cs}) , \tag{4} 
\end{align}

\subsection{Affinity-enhanced attentional network}
\label{sec:32}
The core concept of our AEANet involves passing the feature maps $ (F_c, F_s) $, extracted by a pre-trained VGG-19 network, into the content affinity-enhanced attentional module and style affinity-enhanced attentional module, respectively. Through two independent CAEA and SAEA modules, we perform localized feature extraction and detail enhanced for style and content, gaining $ F_{cc} $ and $ F_{ss} $.  These outputs are subsequently fed into HA module, where they are recombined to produce $ F_{cs} $.

\subsubsection{Content affinity-enhanced attentional}
The content affinity-enhanced attentional module is divided into two steps: 
(1) calculating an attention map by utilizing deep and shallow features of the content image; (2) enhancing the details of the content image.

\textbf{Calculate attention map: }
Generally, Softmax attention can be expressed as.
\begin{equation*}
S = \sigma(QK^T)*V = Attn(Q,K,V), \tag{5}
\end{equation*}
Inspired by agent attention~\cite{han2023agent}, our attention calculation method is:

\begin{equation*}
\begin{split}
S &=  Attn(Q,A,Attn(A,K,V))\\
&= \sigma(QA^T)\sigma(AK^T)V, 
\end{split}\tag{6}
\end{equation*}

$ \sigma (\cdot)$ represents Softmax function. $A$ is a newly defined agent token. In the process of style transfer, it is necessary to preserve the texture and structure of the content image while achieving stylization. Therefore, we designed a content affinity attention module that adaptively captures the 
local texture structures and global statistics of the content image. We denote the content feature map 
as $F_c \in \mathbb{R}^{C \times H \times W}$, which is then fed into three convolutional layers to generate 
three new feature maps: \( F_{cq} \), \( F_{ck} \), and \( F_{cv} \).

\begin{equation*}
F_{cq} = Conv(F_c)
\end{equation*}
\begin{equation*}
F_{ck} = Conv(F_c) \tag{7}
\end{equation*}
\begin{equation*}
F_{cv} = Conv(F_c)
\end{equation*}
Each convolution is a learnable \( 1 \times 1 \) convolution. Than, spatial attention map $A \in \mathbb{R}^{P \times P}$ ,  where $P = W\times H$,
\begin{equation*}
A = \text{Softmax}(F_{cq}^T \otimes F_{ck}) \otimes F_{cv}, \tag{8}
\end{equation*}
\begin{equation*}
F_{aff} = \text{Softmax}(F_{cq}^T \otimes F_{cq}) \otimes A , \tag{9}
\end{equation*}
where $\otimes $ is matrix multiplication. 

\textbf{Content detail-enhanced: }  
We begin by subtracting the square of the attention score matrix \( F_{\text{aff}} \) from itself, passing the result through the ReLU activation function, and then taking the square root of this matrix; This process yields a new matrix---weight matrix $M_{\text{weight}}$, which maintains the original matrix's shape and has values distributed between 0 and 1. Subsequently, we standardize \( F_{\text{cv}} \), multiply it by the detailed weight matrix $M_{\text{weight}}$, and finally add it to the original \( F_{\text{aff}} \) matrix.The specific formula is as follows: 

\begin{equation}
    M_{\text{weight}} = \sqrt{\text{ReLU}\left(F_{\text{aff}} - F_{\text{aff}}^2\right)} \tag{10}
\end{equation}

\begin{equation}
    F_{\text{cc}} = M_{\text{weight}}  \times Norm \left(F_{\text{cv}}\right) + F_{\text{aff}} \tag{11}
\end{equation}

Where ReLU represents the activation function and $ Norm $ denotes mean-variance normalization. It is worth noting that the design of \textbf{the SAEA module is the same as that of the CAEA module.}

\subsubsection{Hybrid attentional}

The CAEA and SAEA modules process the content and style images, respectively, to generate the 
attention feature maps \( F_{\text{cc}} \) and \( F_{\text{ss}} \). Each feature map undergoes mean-variance normalization and  \( 1 \times 1 \) convolution layer. Subsequently, Softmax operation is applied to the 
product of \( F_{\text{cc}} \) and \( F_{\text{ss}} \), which is then matrix-multiplied with \( F_{\text{ss}} \) to 
produce \( F_{\text{cs}}' \). Adding \( F_{\text{cc}} \) to \( F_{\text{cs}}' \) results in the final stylized 
feature map \( F_{\text{cs}} \). The HA module adjusts the style features based on the distribution of the content 
features, thereby better capturing the mapping relationship between content and style features. The specific formula is as follows: 

\begin{equation}
    F_{\text{cs}} = F_{\text{ss2}} \otimes \text{SoftMax}\left(F_{\text{cc1}}^T \otimes F_{\text{ss1}}\right) + F_{\text{cc}}, \tag{12}
\end{equation}

where $\otimes$ denotes matrix multiplication.

\subsection{Loss function}
\label{sec:33}
It is well known that a network model's loss function is its ultimate optimization objective.
In our network model, the loss functions utilized include the content loss $ L_c $, the local dissimilarity content loss $ L_{LD\_Content} $, the style loss $ L_{s}$, the local dissimilarity style loss $ L_{LD\_Style}$, and the identity loss $ L_{identity} $. Specifically:

\begin{equation*}
\begin{split}
    \mathcal{L}oss = \lambda_c \mathcal{L}_{c}^{i} + \lambda_s \sum_{j=1}^{N} \mathcal{L}_s + \lambda_{id} \mathcal{L}_{identity}    \\
    + \lambda_{cld} \mathcal{L}_{LD\_Content}^{i} + \lambda_{sld} \sum_{j=1}^{N} \mathcal{L}_{LD\_Style}^{j}, 
\end{split}\tag{13}
\end{equation*}

\textbf{Perceptual loss:}
Similar to AdaIN~\cite{huang2017arbitrary}, we minimize the loss \( L_c \) between the content image \( I_c \) and the stylized image \( I_{cs} \), as well as the loss \( L_s \) between the style image \( I_s \) and \( I_{cs} \).
\begin{equation}
    \mathcal{L}_c = \sum_{i=1}^{L} \left\| \phi_i(I_c) - \phi_i(I_{cs}) \right\|_2, \tag{14}
    \label{eq:contentloss}
\end{equation}
\begin{equation*}
\begin{split}
    \mathcal{L}_{\text{s}}^i = \left\| \mu(\phi_i(I_{cs})) - \mu(\phi_i(I_c)) \right\|_2 + \\
    \left\| \sigma(\phi_i(I_{cs})) - \sigma(\phi_i(I_c)) \right\|_2,
\end{split}\tag{15}
\label{eq:styleloss}
\end{equation*}
\begin{figure*}[th]
\begin{center}
    \centering
    \includegraphics[width=1\textwidth]{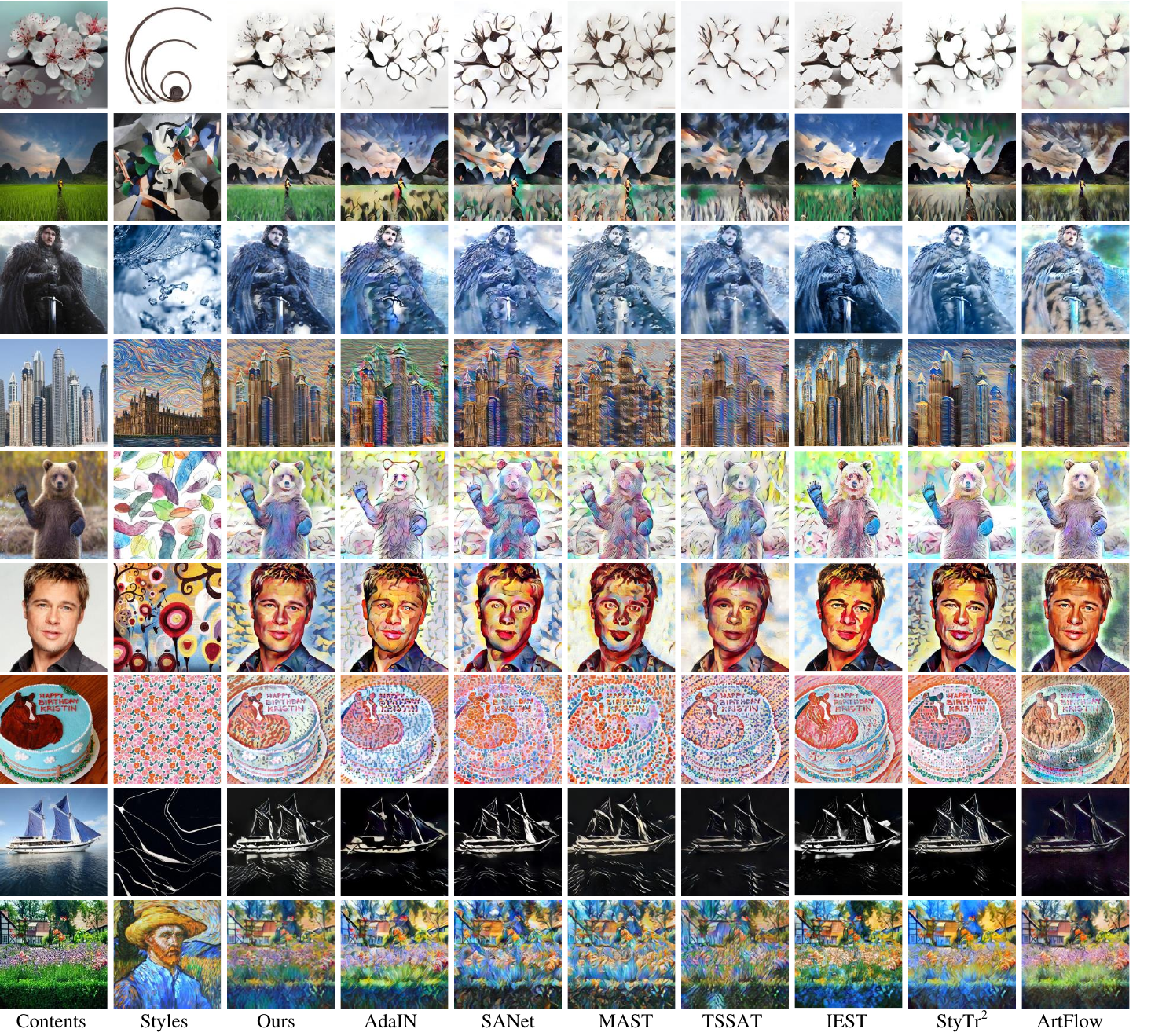}
\end{center}
   \caption{ Qualitative comparisons of style transfer results with SOTA methods }
\label{fig:compare}
\end{figure*}
Where \textbf{\( \mu(\cdot) \) }denotes the mean of the features, \( \phi \)  represents the features extracted from the \( i \)th layer of the pre-trained VGG-19 network, and \textbf{\( \sigma(\cdot) \)} denotes the variance of the features.

\textbf{Identity loss: }
follow SANet~\cite{park2019arbitrary}, We introduce identity loss to better preserve both content and style features.

\begin{equation*}
\begin{split}
    \mathcal{L}_{identity} = \lambda_{id1} (\left\| I_{cc} - I_c \right\|_2 +  \left\| I_{ss} - I_s \right\|_2) + \\
     \lambda_{id2}\sum_{i=1}^{L} \left( \left\| \phi_i(I_{cc}) - \phi_i(I_c) \right\|_2 +  \left\| \phi_i(I_{ss}) - \phi_i(I_s) \right\|_2 \right), 
\end{split} \tag{16} \label{eq:identity_loss}
\end{equation*}

Where \( I_{cc} \) and \( I_{ss} \) represent the composite results generated when the inputs are identical content and style images, respectively. \(\phi\) refers to the features extracted from the \(relui\_1(i \in[1-5])\) layers of the VGG network, while \(\lambda_{id1}\) and \(\lambda_{id2}\) are hyperparameters.

\textbf{Local dissimilarity loss:} During style transfer, when generating stylized images using the same content image and multiple style images, local dissimilarity loss is employed to compare further the difference between the features processed by AEANet and those after random permutation of the features. This ensures that the model learns content and style features and exhibits a degree of robustness to minor variations in the input. This is achieved by randomly permuting the batch dimension of the $Ics1$ tensor and then computing the content loss between the original features and those after random permutation. The content loss is calculated by comparing the higher-order feature maps of the two feature tensors. 
\begin{table*}
\begin{center}
\begin{tabular}{|l|c|c|c|c|c|c|c|c|}
\hline
Model & Ours & AdaIN  & SANet & MAST & TSSAT & IEST & StyTr$^2$ & ArtFlow  \\
\hline\hline
$L_c \downarrow $ & \textbf{1.96}  & 2.34  & 2.44 & 2.46& 3.34  & 1.97 & \underline{1.91} & 2.13\\
$L_s \downarrow $ & 1.62 & 1.91  & \underline{1.18} & 1.55 & 1.92  & 3.47 & \textbf{1.47} & 3.08\\
$SSIM \uparrow $ & \textbf{0.47} & 0.37 & 0.37 & 0.38& 0.43 & \textbf{0.47} & \underline{0.57} & 0.46\\
$Time/sec \downarrow $ & 0.045 & \underline{0.011} & 0.016 & 0.022 & 0.128  & \textbf{0.013} & 0.165 & 0.093\\
\hline
\end{tabular}
\end{center}
\caption{Quantitative evaluation. Comparison of style transfer models based on $L_c$ and $L_s$ performance metrics,
we evaluate the effectiveness of various methods in preserving input content and style by calculating the average content and style loss values. SSIM~\cite{wang2004image} is a metric used to measure the similarity between two images, in style transfer, a higher value indicates a higher degree of similarity, we calculate the average SSIM using the results of stylization generated from the same content image and 60 different style photos. We use $512 \times 512$ pixel images to calculate the average generation time, we use the average time to generate 300 stylized photos. The $ \uparrow $ indicates that a higher value signifies better performance, whereas a  $\downarrow $ indicates that a lower value is better. The best results are \underline{underlined}, and the second best is in \textbf{bold}.
}
\label{tab:comparetable}
\end{table*}
The calculation of the style loss between the original style features and the permuted style features typically involves comparing multiple layers of feature maps (from $ Ics1feats[0] $ to $Ics1feats[4]$) to assess the consistency of the style comprehensively. The local dissimilarity loss enables the model to learn content and style features with a degree of affinity, maintaining consistency with the content and style images even when the order of the input data changes.

\begin{equation*}
    \mathcal{L}_{LD\_Content}^i = \left\| \phi_i(I_{cs1}) - \phi_i(I_{cs2}) \right\|_2, \tag{17}
\end{equation*}

\begin{equation*}
\begin{split}
    \mathcal{L}_{LD\_Style}^i =  \sum_{i=1}^{L} ( \left\| \mu(\phi_i(I_{sc1})) - \mu(\phi_i(I_{sc2})) \right\|_2 + \\
    \left\| \sigma(\phi_i(I_{sc1})) - \sigma(\phi_i(I_{sc2})) \right\|_2),
\end{split}\tag{18}
\end{equation*}

where \( I_{cs1} \) and \( I_{cs2} \) represent the images generated from the same content image but with different style images, while \( I_{sc1} \) and \( I_{sc2} \) represent the images generated from the same style image but with different content images.

\section{Experiments}

\subsection{Implementation details} 

As described in Section \ref{sec:3}, our model primarily consists of an encoder $ E $, a decoder $ D $, and AEANet, as illustrated in Figure~\ref{fig:model}. The encoder comprises a pre-trained VGG-19 network~\cite{2014Very}, with the encoder and decoder being symmetrically designed. The VGG-19 layers conv1\_1, conv2\_1, conv3\_1, and conv4\_1 extract features from the input style and content images. The input to the content affinity attention module is the feature extracted from the conv4\_1  layer of VGG-19. The model is iterated 160,000 times on an NVIDIA A100 Tensor Core GPU. We employed the Adam optimizer for training, with a style image weight of 5.0, content image weight of 1.0, batch size of 8, and learning rate of 0.0001.

\textbf{Datasets: }We used MS-COCO~\cite{Lin2015Microsoft} as the content dataset and WikiArt~\cite{phillips2011wiki} as the style dataset. During training, input images were loaded at a resolution of $512 \times 512$ pixels and further randomly cropped to $ 256 \times 256 $ pixels to implement data augmentation strategies. During inference, our model can handle content and style images of arbitrary resolutions.

\textbf{Baselines: }We compared our model against several state-of-the-art (SOTA) style transfer models, including AdaIN\cite{huang2017arbitrary},SANet~\cite{park2019arbitrary},MAST~\cite{deng2020arbitrary},TSSAT~\cite{chen2023tssat},IEST~\cite{chen2021artistic}, StyTr$^2$~\cite{deng2022stytr2}, and ArtFlow~\cite{an2021artflow}. For these SOTA methods, we used their publicly available code and models to infer style transfer results.

\textbf{Timing information: }Our model was trained on an NVIDIA A100 Tensor Core GPU for approximately 18 hours. Table ~\ref{tab:comparetable} compares the inference time for generating $512 \times 512$ resolution images with other SOTA methods.

\subsection{Comparison with State-of-the-Art methods}

\textbf{Qualitative evaluation: }Figure \ref{fig:compare} presents a visual comparison of our results with eight different SOTA models. It can be observed that AdaIN~\cite{huang2017arbitrary} captures less information from the style image due to the simplified alignment of mean and variance, resulting in the presence of colors in the generated image that are not found in either the content or style images (e.g.,$4^{th}$,$5^{th}$ and $6^{th}$rows). SANet~\cite{park2019arbitrary} excessively introduces semantic structures from the style image (e.g.,$3^{th}$,$5^{th}$,$7^{th}$ and $9^{th}$ rows). Although MAST~\cite{deng2020arbitrary} employs an attention mechanism, its application does not yield the expected results. Analyses indicate that while the main structure of the content image remains clear, other structural parts are compromised. The style features in the generated images are insufficiently represented (e.g.,$1^{th}$,$2^{th}$,$5^{th}$ and $8^{th}$ rows). For TSSAT~\cite{chen2023tssat}, despite balancing global statistics and local features, the preservation of the content image structure is suboptimal, resulting in some unappealing outcomes (e.g.,$1^{th}$,$3^{th}$,$8^{th}$ and $9^{th}$ rows). IEST~\cite{chen2021artistic} extracts fewer features from the content image, which leads to subpar preservation of the content structure in the stylized images (e.g.,$1^{th}$,$3^{th}$ and $4^{th}$ rows). The StyTr$^2$~\cite{deng2022stytr2} method replaces the traditional VGG~\cite{2014Very} encoder with a Transformer~\cite{vaswani2017attention} encoder, which mitigates the drawbacks of reduced receptive fields during the encoding phase. However, content image information loss is still evident (e.g.,$1^{th}$,$2^{th}$ and $8^{th}$ rows). The ArtFlow~\cite{an2021artflow} 
the model produces stylized images during inference that exhibit noticeable deviations from the input style images 
(e.g.,$1^{th}$,$3^{th}$,$7^{th}$ and $9^{th}$ rows). Consequently, our research findings demonstrate an 
effective preservation of content structure integrity and achieve desirable expression of style features.

\textbf{Quantitative evaluation:} We evaluate the quality of style transfer by calculating the content discrepancy between the generated images and the input content images and the style discrepancy between the generated images and 
\begin{figure*}[th]
\begin{center}
    \centering
    \includegraphics[width=1\textwidth]{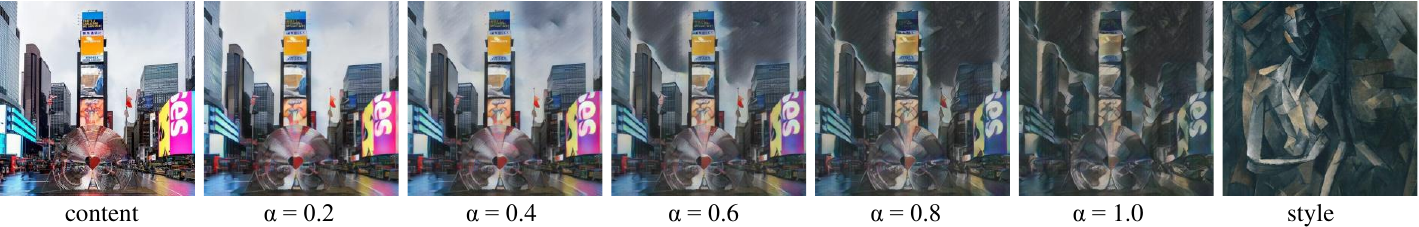}
\end{center}
   \caption{ content style trade-off }
\label{fig:tradeoff}
\end{figure*}
\begin{figure}[th]
\begin{center}
    \centering
    \includegraphics[width=0.5\textwidth]{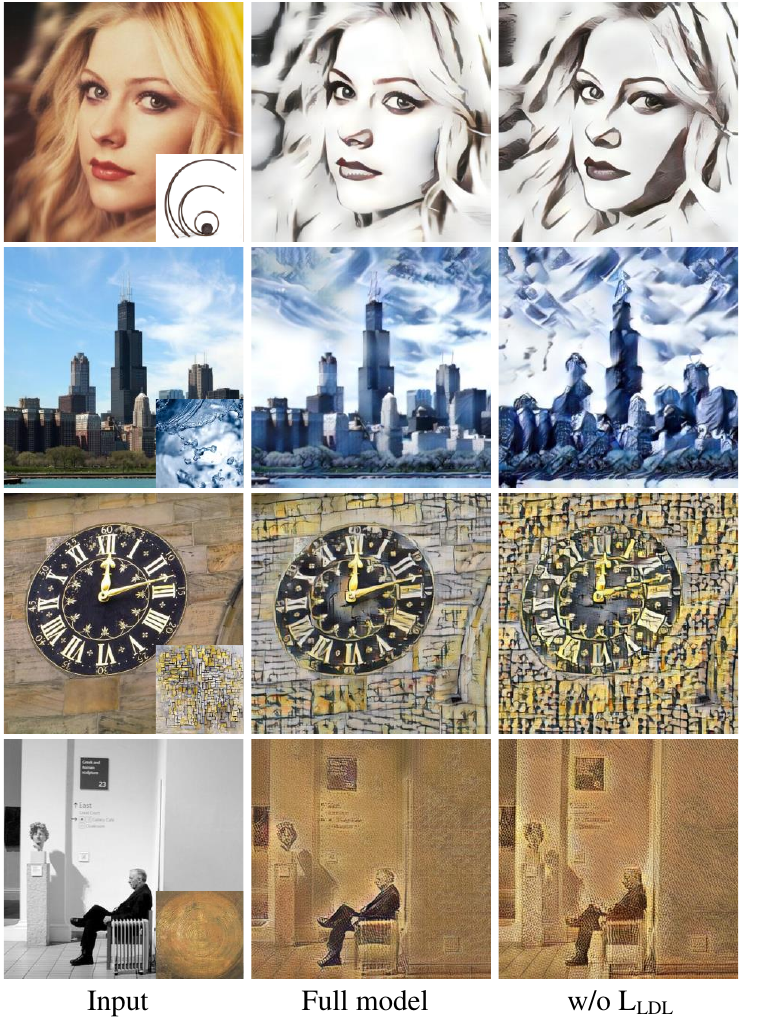}
\end{center}
   \caption{ Ablation study results. The first column shows the input content and style images, the second column shows the stylization results from the complete model, and the third column shows the results without the  $ L_{LDL} $ loss functions ($ L_{LDL} $ is $\mathcal{L}_{LD\_Content} $ add 
$ \mathcal{L}_{LD\_Style} $ ). }
\label{fig:ablation}
\end{figure}
the input style images. These two metrics serve as indirect measures. Intuitively, more minor discrepancies indicate better preservation of content and style. Content discrepancy is computed using Equation \ref{eq:contentloss}, and style discrepancy is calculated using Equation \ref{eq:styleloss}.  Table~\ref{tab:comparetable} presents the corresponding quantitative results. The results show that our method achieves an optimal balance between content preservation and style transfer, which is validated by the intuitive comparison results in Figure~\ref{fig:compare}. Additionally, we introduce the Structural Similarity Index (SSIM) ~\cite{wang2004image}parameter in Table~\ref{tab:comparetable}. SSIM is a metric used to evaluate the visual similarity between two images, particularly when comparing two different versions of an image.

\subsection{Ablation study}
\textbf{Local dissimilarity loss:} To validate the effectiveness of our loss function, we removed the $ \mathcal{L}_{LD\_Style} $ and $ \mathcal{L}_{LD\_Content} $ loss from the model. As shown in Figure~\ref{fig:ablation}, compared with the model results without the  $\mathcal{L}_{LD\_Style}$ and $\mathcal{L}_{LD\_Content}$ loss function, the stylized images reveal that without $\mathcal{L}_{LD\_Style}$ and $\mathcal{L}_{LD\_Content}$, there are many redundant texture features from the style image, and the preservation of content image features is also not significant.

\subsection{Applications}
\textbf{Content–style trade-off:} During model inference, we can adjust the value of $ \alpha $ in the following function to control the degree of stylization in the image:

\begin{equation*}
    F_{cs} = \text{Decoder}(\alpha F_{cs} + (1-\alpha)F_c), \tag{19}
\end{equation*}

When \( \alpha = 0 \), the result is a non-stylized image, and when \( \alpha = 1 \), the result is a fully stylized image. We have included the stylization results for varying values of \( \alpha \) from 0 to 1 in Figure~\ref{fig:tradeoff}.

\section{ Conclusion }

In this paper, we propose AEANet to achieve high-quality style transfer. Initially, the model performs a local style feature information exchange and detail enhanced on both the content and style images. The HA module adjusts the style features based on the distribution of the fine-tuned content features. Our model achieves a good balance between preserving the structural texture of the content image and stylization. Experimental results demonstrate that our network fully considers local content and style features. The generated results attest to the superiority of our model. The main limitation of our study is the room for improvement in the output speed of high-resolution images. In future work, we aim to simplify computational complexity to enhance output efficiency.


{\small
\bibliographystyle{cvm}
\bibliography{cvmbib}
}

\end{document}